\documentclass[letterpaper]{article} 
\usepackage{aaai2026}  
\usepackage{times}  
\usepackage{helvet}  
\usepackage{courier}  
\usepackage[hyphens]{url}  
\usepackage{graphicx} 
\usepackage{natbib}  
\usepackage{caption} 
\usepackage{algorithm}
\usepackage{algorithmic}

\usepackage{newfloat}
\usepackage{listings}
\DeclareCaptionStyle{ruled}{labelfont=normalfont,labelsep=colon,strut=off} 
\floatstyle{ruled}
\newfloat{listing}{tb}{lst}{}
\floatname{listing}{Listing}
\usepackage{makecell, tabularx, booktabs}
\usepackage{multirow, multicol, amsmath, amssymb}
\usepackage{adjustbox}
\usepackage{subcaption}
\usepackage{tcolorbox}

\tcbset{
  promptbox/.style={
    colback=gray!10,    
    colframe=gray!50,   
    boxrule=0.5pt,        
    arc=3mm,            
    boxsep=1mm
  }
}

\title{Next Generation Active Learning: Mixture of LLMs in the Loop}
\author{
    Yuanyuan Qi\textsuperscript{\rm 1}, Xiaohao Yang\textsuperscript{\rm 1}, Jueqing Lu\textsuperscript{\rm 1}, Guoxiang Guo\textsuperscript{\rm 1}\\
    Joanne Enticott\textsuperscript{\rm 1}, Gang Liu\textsuperscript{\rm 2}\thanks{Corresponding author}, Lan Du\textsuperscript{\rm 1}
}
\affiliations{
    \textsuperscript{\rm 1}Monash University\\
    \textsuperscript{\rm 2}	Harbin Engineering University\\
    \{yuanyuan.qi, xiaohao.yang, jueqing.lu, guoxiang.guo, joanne.enticott, lan.du\}@monash.edu,
    liugang@hrbeu.edu.cn
}

\begin{document}
\maketitle
\begin{abstract}
With the rapid advancement and strong generalization capabilities of large language models (LLMs), 
they have been increasingly incorporated into the active learning pipelines as annotators to reduce annotation costs. 
However, considering the annotation quality, labels generated by LLMs often fall short of real-world applicability.
To address this, we propose a novel active learning framework, Mixture of LLMs in the Loop Active Learning, 
replacing human annotators with labels generated through a Mixture-of-LLMs-based annotation model, aimed at enhancing LLM-based annotation robustness by aggregating the strengths of multiple LLMs. 
To further mitigate the impact of the noisy labels, 
we introduce annotation discrepancy and negative learning to identify the unreliable annotations and enhance learning effectiveness.
Extensive experiments demonstrate that our framework achieves performance comparable to human annotation and consistently outperforms single-LLM baselines and other LLM-ensemble-based approaches. 
Moreover, our framework is built on lightweight LLMs, enabling it to operate fully on local machines in real-world applications.
\end{abstract}

\begin{links}
    \link{Code}{https://github.com/qijindou/MoLLIA}
    \link{Appendix}{https://github.com/qijindou/MoLLIA/tree/main/Appx}
\end{links}

\section{Introduction}
Active Learning (AL) is a paradigm in machine learning that strategically selects informative samples for annotation, with the objective of minimizing labeling costs while achieving predictive performance comparable to models trained on fully labeled datasets \cite{ren2021survey, wu2025alscope, werner2024cross}.
With the rapid advancement of Large Language Models (LLMs) \cite{brown2020language}, 
renowned for their remarkable generalization capabilities \cite{openai2023gpt4}, 
these models have increasingly been integrated into the conventional active learning workflow, 
enhancing the cost-efficiency of the annotation process \cite{kholodna2024llms}, 
and marking a significant evolution in the active learning landscape \cite{xia2025selection}.

The integration of LLMs into the active learning workflow offers a promising route to improving annotation efficiency, however, their reliability as annotators remains an open challenge \cite{ding2023gpt, ming2024autolabel}. 
Since LLMs are trained for general purpose tasks, 
their performance often degrades due to domain shift when applied to specialized datasets, 
resulting in annotation quality that is typically insufficient to serve as oracle labels \cite{gligoric2024can, guo2024mortar}. 
LLM-ensemble methods, which combine multiple LLMs with diverse architectures or training paradigms, 
offer a more robust and effective approach to enhancing annotation precision \cite{chen2025harnessing}.
Building upon this idea, leveraging the outputs from a Mixture of LLMs as annotators offers a feasible and more reliable strategy for transforming traditional human in the loop active learning into Mixture of LLMs in the loop active learning.
Moreover, the active learning process can operate in a semi-fully, or even fully, human-free manner, while maintaining a relatively high standard of annotation quality.

\begin{figure}
    \centering
    \includegraphics[width=\columnwidth]{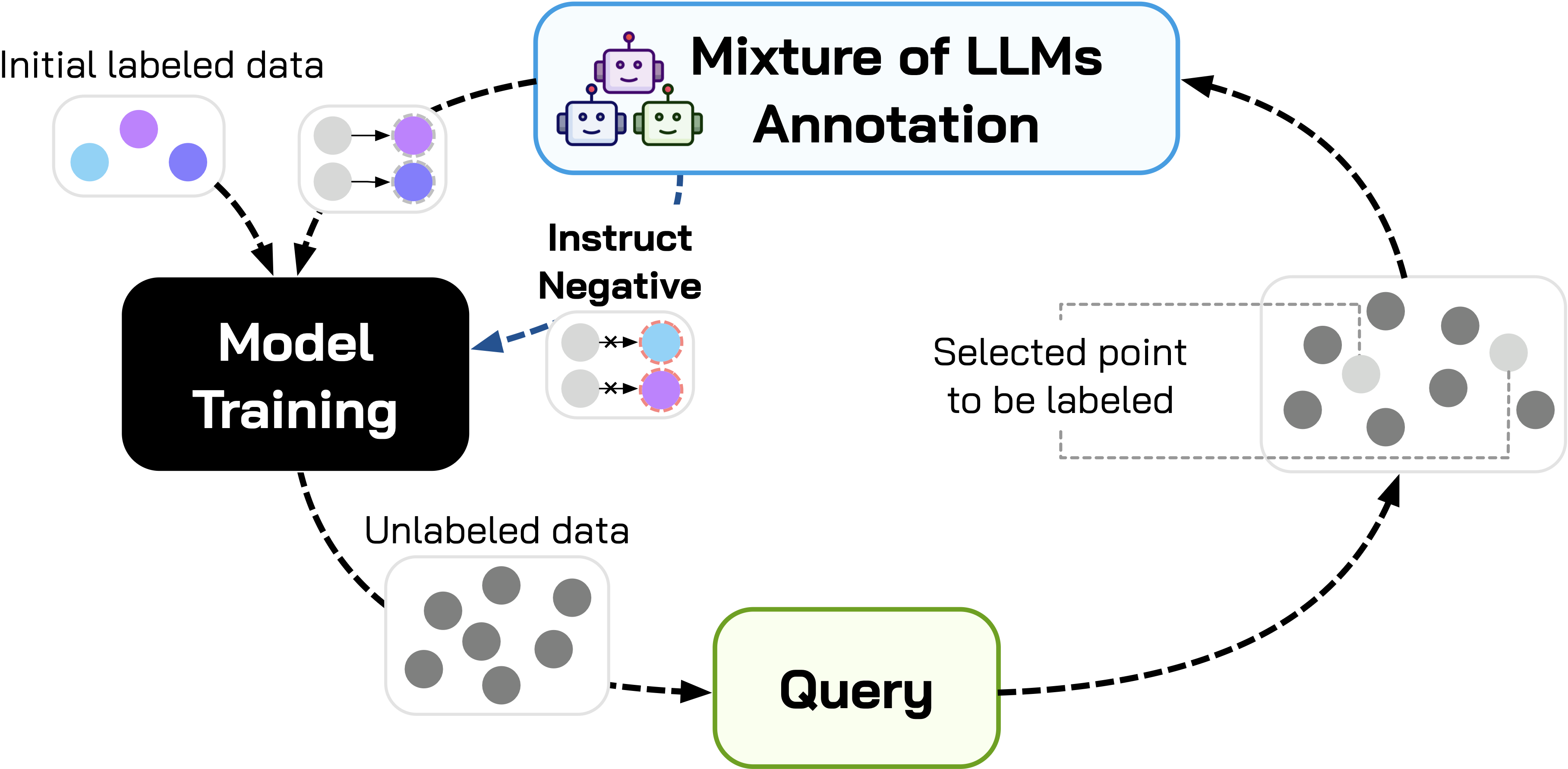} 
    \caption{Overview of MoLLIA.}
    \label{fig:overview}
\end{figure}

Although model performance typically scales with the volume of training data, 
real-world applications often suffer from a shortage of labeled data. 
In active learning settings, this limitation is even more pronounced, as usually only a small subset of ground truth labeled data is available at the initial iteration.
Consequently, fully exploiting the limited labeled data to enhance the quality of Mixture of LLMs generated annotations is critical for the success of Mixture of LLMs in the loop active learning. 
Additionally, since the generated labels may be noisy or unreliable \cite{yang2024neural}, 
using them directly as supervision labels can degrade model performance and compromise the reliability of the active learning process \cite{lu2025navigating}. 
To mitigate this issue, adopting noisy label learning techniques that compensate for label noise has proven to be a promising solution for handling unreliable annotations \cite{song2022learning}. 
Besides, regarding the AL model as a task specific small language model (SLM) which is trained or finetuned on the target dataset, 
the mismatch between the SLM and LLM-based annotators for annotation provides an additional insights of annotation discrepancy. 
And this discrepancy can be exploited to further refine the annotation process \cite{yuan2024hide}. 
Integrating these strategies into the Mixture of LLMs framework allows the AL model to remain robust in the presence of imperfect annotations, 
thereby boosting both annotation quality and downstream model performance.

Overall, to further reduce human labeling costs while maintaining the reliability of active learning process for practical deployment, 
we propose the \textbf{M}ixture \textbf{o}f \textbf{L}\textbf{L}Ms \textbf{I}n the Loop \textbf{A}ctive Learning (MoLLIA) framework (Fig.~\ref{fig:overview}). Specifically, at each active learning iteration, we select samples for annotation using existing acquisition strategies, and employ multiple lightweight LLMs to generate candidate labels. These outputs are then aggregated through a mixture module to determine the final annotation used in the next iteration training.
To mitigate the effects of noisy labels, we further incorporate negative learning alongside annotation discrepancy between AL model and LLMs.
In summary, our MLAL framework offers the following key contributions:

\begin{itemize}
    \item \textbf{Human-Free Active Learning}: 
    We propose a novel zero-human annotation active learning framework based on a Mixture-of-LLMs-based annotation model (MoLAM). By aggregating MoLAM as annotators and incorporating other learning mechanism, our method achieves annotation-free active learning with performance comparable to traditional human in the loop approaches.
    \item \textbf{Robust Active Learning}:
    We leverage the disagreement between the AL model (treated as a task specific SLM) and the LLM-based annotator as an annotation discrepancy indicator and incorporate a negative learning mechanism to improve the robustness of the learning process.
    \item \textbf{Reliable Empirical Validation}:
    We validate the effectiveness of proposed framework across four widely used benchmark datasets and multiple active learning strategies. MoLLIA achieves superior performance and demonstrates comparability to human annotators.
\end{itemize}

\section{Related Work}
With the rapid advancement of LLMs, their integration into active learning has become increasingly prevalent. 
Traditional AL methods rely on carefully designed uncertainty metrics and sample selection strategies to maximize model performance while minimizing annotation costs \cite{ren2021survey, werner2024cross, qi2024multi}. 
Recently, due to their strong generalization capabilities and extensive inherent knowledge, 
LLMs have been incorporated into AL pipelines, either in the sampling or annotation stages, to further reduce labeling costs \cite{azeemi2024language, xia2025selection}.
To utilize LLMs as annotators, \citeauthor{kholodna2024llms} \shortcite{kholodna2024llms} employ inter-annotator agreement to evaluate the consistency of multiple LLMs and select the most reliable one to replace human annotators. 
\citeauthor{rouzegar2024enhancing} \shortcite{rouzegar2024enhancing} propose a hybrid annotation framework that combines LLM-generated labels with human annotations based on LLMs uncertainty. 
However, due to the limited quality of LLM-generated labels, 
these approaches either fail to match the performance of oracle labels or still require substantial human annotation effort.
While methods such as NoisyAL \cite{yuan2024hide} and FreeAL\cite{xiao2023freeal} incorporate both LLMs and smaller models to generate and refine labels, 
they are fundamentally noisy supervised learning approaches rather than active learning frameworks, as they lack iterative sample selection guided by trainable-model uncertainty. 
In addition, most existing methods depend on commercial API calls, 
raising unresolved concerns about data privacy and security, particularly in sensitive or real-world applications.

To ensure the reliability and effectiveness of LLM-generated outputs, 
numerous studies have explored techniques for estimating their quality, 
with a primary focus on uncertainty estimation.
Overall, uncertainty estimation in LLMs can be broadly categorized into three main approaches: verbalization-based, consistency-based, and logit-based.
Verbalization-based methods rely on prompting LLMs to self-assess its confidence by explicitly asking for likelihood judgments or uncertainty estimates through natural language responses \cite{yona2024can, linteaching}.
Consistency-based methods estimate uncertainty by generating multiple responses for the same input and analyzing their variability \cite{chen2024quantifying, tian2023just}.
Logit-based methods derive uncertainty from the model’s internal probability distribution, 
using metrics such as entropy or margin over predicted tokens to quantify confidence \cite{kuhnsemantic, abbasi2024believe, zhang2025dpcore}.

While uncertainty estimation offers valuable insights into the confidence of LLM predictions, it does not directly address the quality of the final annotations. 
To further enhance annotation reliability, 
LLM-ensemble methods have emerged as a widely adopted strategy that leverages the complementary strengths of multiple LLMs \cite{chen2025harnessing}. 
Recent advances in this area can be broadly categorized into two groups: 
consensus-oriented and diversity-oriented approaches. 
Consensus-oriented methods aim to select the output that exhibits the highest agreement across multiple responses, often relying on voting or similarity metrics \cite{li2024more, guha2024smoothie, si2023getting}. 
In contrast, diversity-oriented methods focus on analyzing of differences across candidate outputs to resolve conflicts or synthesize more informative and robust responses\cite{jiang2023llm, tekin2024llm, lv2024urg}.
However, given that label generation lacks semantic structure in the output space, consensus-oriented approaches are more suitable for our task, 
as they align better with the discrete and bounded nature of classification labels.

\begin{figure*}
    \centering
    \includegraphics[width=0.85\textwidth]{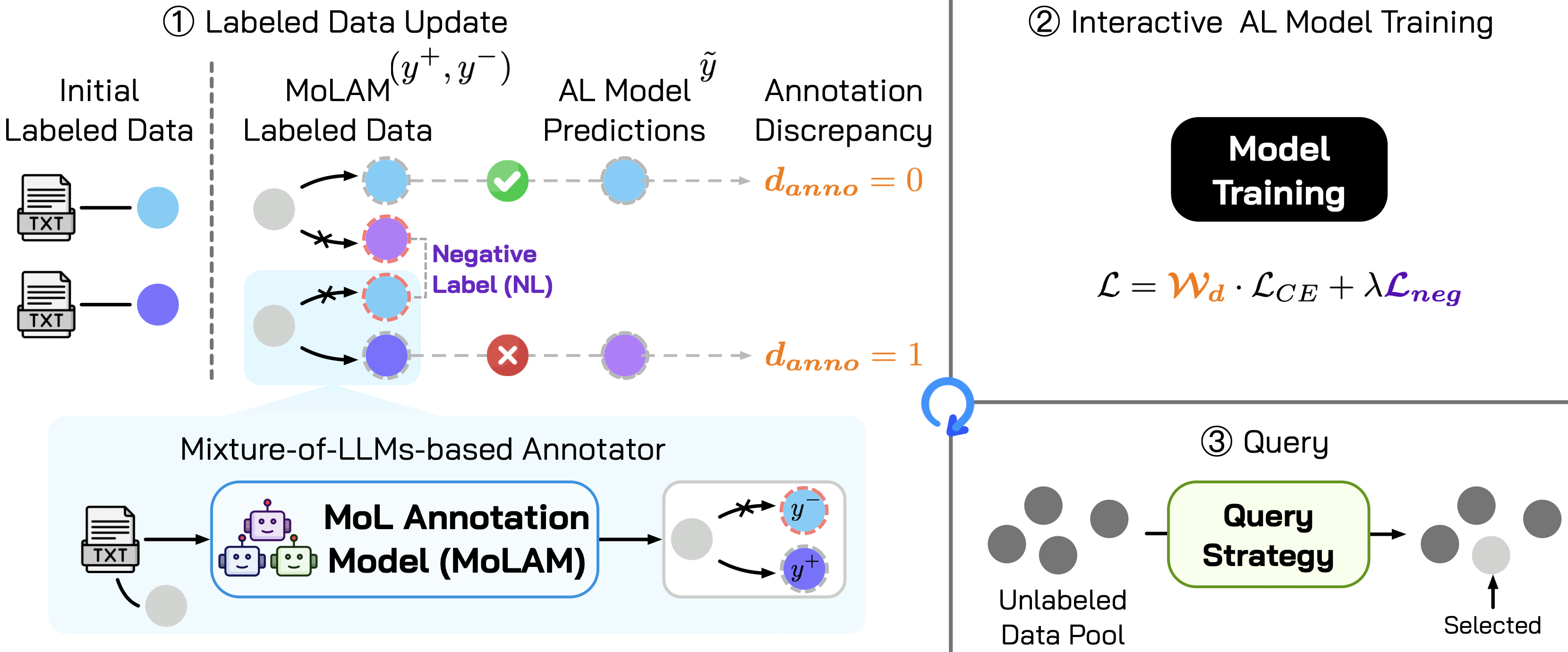} 
    \caption{Workflow of MoLLIA framework. The AL model is first trained on the initial labeled dataset and used to query the most informative instances for annotation. A Mixture-of-LLMs-based annotator then generates labels $y^+$ and corresponding negative labels $y^-$ for the selected instances (as detailed in Mixture-of-LLMs-based Annotation Model section). The annotation discrepancy ($d_{anno}$) is computed based on the disagreement between the AL model’s predictions and the Mixture-of-LLMs-based annotator. The AL model is then updated using a loss function that incorporates both weighted annotation discrepancy and negative learning, and the querying process is iteratively repeated based on the updated AL model.}
    \label{fig:workflow}
\end{figure*}

\section{Methodology}
Without loss of generality, let $L=\{X,Y\}$, $U=\{X\}$ represent the initial collection of training set and unlabeled data samples, where $|U| \gg |L|$. Here, $Y \subset \{1,\ldots,K\}$ denotes the set of multi-class labels, and $K$ is the total number of classes. 
Fig.~\ref{fig:workflow} illustrates the workflow of our proposed framework, MoLLIA.
Our method adopts standard active learning query strategies but replaces human annotation with the Mixture-of-LLMs-based Annotation Model (MoLAM). 
While MoLAM improves annotation quality compared to single LLM annotators, its outputs may still include noisy labels and therefore remain inferior to human level annotation quality.
To further mitigate this effects, we propose Robust Active Learning, which enhances robustness through two key mechanisms.
The first part utilize the negative labels, classes that an instance is unlikely to belong to, provided by MoLAM to guide the AL model away from incorrect predictions, thereby improving learning efficiency and class discrimination.
The second part leverage the annotation discrepancy, quantifying the disagreement between MoLAM predicted labels and the AL model’s predictions, to re-weight the training loss, reducing the influence of potentially incorrect annotations.

\subsection{Mixture-of-LLMs-based Annotation Model}
Instead of relying on human annotators, MoLLIA further reduces annotation costs by introducing a fully human-free annotation model, MoLAM. The core idea of MoLAM is that LLMs with different architectures exhibit varying performance across datasets \cite{jiang2023llm}. Therefore, by aggregating the outputs of several lightweight LLMs, MoLAM generates more reliable and comprehensive labels, delivering annotation quality that is acceptable for downstream active learning training.

\begin{figure}
    \centering
    \includegraphics[width=\columnwidth]{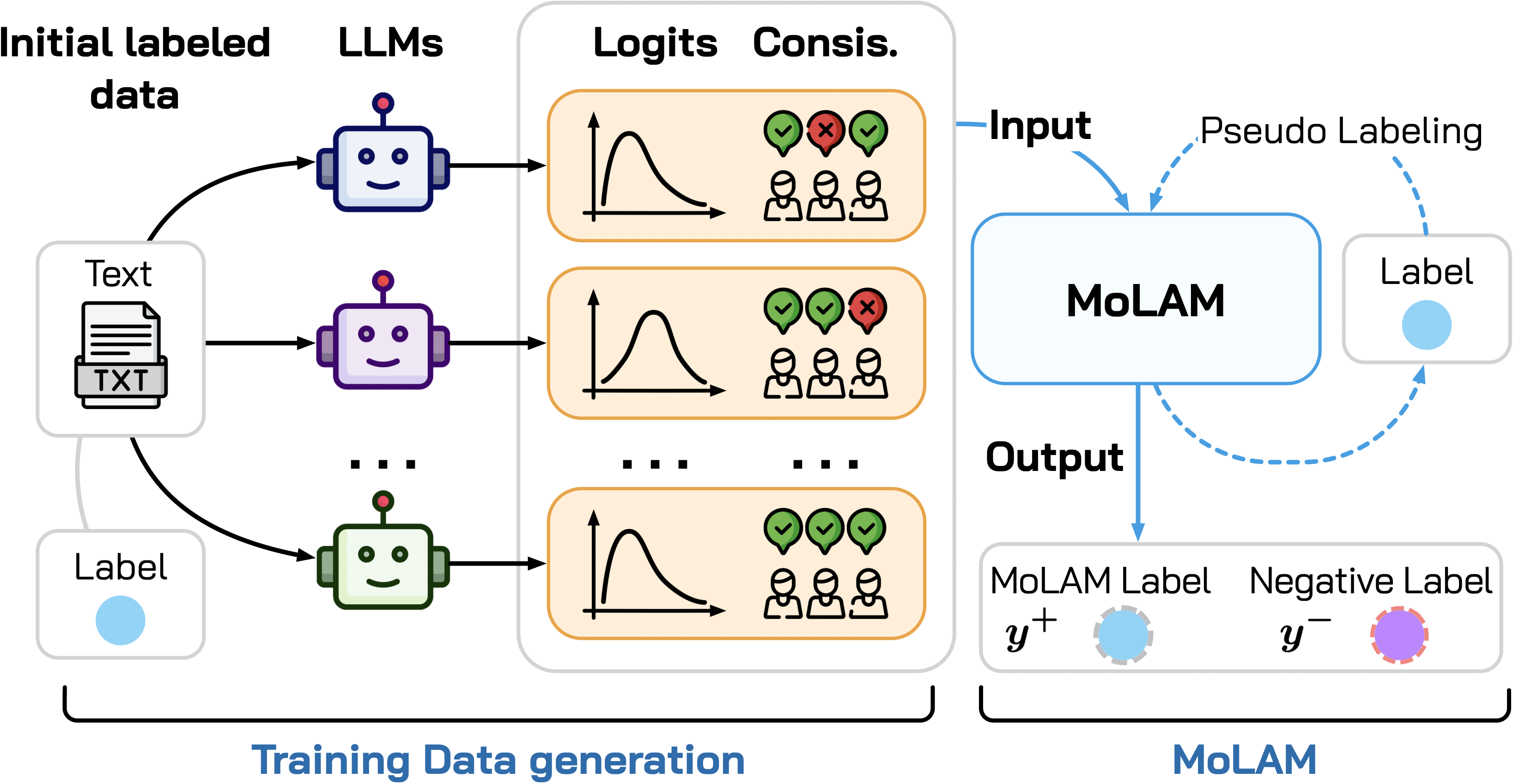} 
    \caption{Overview of MoLAM.}
    \label{fig:molam}
\end{figure}

Figure.~\ref{fig:molam} illustrates the training process of the MoLAM, including training data generation and model training. 
MoLAM is trained solely on the initial labeled dataset, which comprises a very small portion of the entire data pool (only 50 instances). 
Let $\{x,y\}$ denotes one labeled example from the initial labeled set $L$, and let $\mathcal{M}_i\in\{\mathcal{M}_1,...,\mathcal{M}_N\}$ represents the $i_{th}$ LLM among the $N$ LLMs involved in MoLAM.
To construct the training data for MoLAM, each LLM $\mathcal{M}_i$ is queried $T$ times on the same input $x$ to produce: a logits vector, $z_i \in \mathbb{R}^K$, representing the model’s confidence over $K$ candidate labels; a consistency score, $c_i \in \mathbb{R}^K$, where each component $c^{(k)}_i$ indicates how frequently class $k\in\{1,\ldots,K\}$ is predicted across $T$ generations. The computation is formalized as follows:
\begin{align}
    &z_i = \mathcal{M}_i(x),\  z_i \in \mathbb{R}^K \label{eq1} \\
    &\hat{y}_i^t = \text{Decode}(\mathcal{M}_i(x)), \  t=1,\ldots,T \label{eq2} \\
    &c^{(k)}_i = \textstyle\frac{1}{T}\textstyle\sum^T_{t=1}\mathbb{I}[\hat{y}_i^t=k], \  \forall k \in {1, \ldots, K} \label{eq3}
\end{align}

We then train MoLAM using the labeled data in the form of $\{[z_1,c_1,\ldots,z_N,c_N], y\}$, where each input consists of logits and consistency scores obtained from multiple LLMs. Given the limited amount of labeled data available for supervision, we adopt a semi-supervised learning strategy based on a pseudo-labeling mechanism to leverage additional information from the unlabeled pool. 
A confidence threshold $\sigma$ is applied to determine whether an unlabeled instance is reliable enough to be used for training. In this way, MoLAM encapsulates the collective knowledge of Mixture of LLMs and generates the refined label $y^+$, which are then used in subsequent active learning iterations.
Additionally, to further exploit the expert knowledge implicitly encoded in the LLMs, MoLAM identifies negative labels $y^-$,  defined as labels assigned consistently low probabilities (below threshold $\delta$) by all LLMs. These negative labels are integrated into the training process via negative learning to improve the robustness and discriminative ability of the AL model.
The overall procedure is formalized in the following equations:
\begin{align}
    \mathbf{h}(x) &= [z_1,c_1,\ldots,z_N,c_N] \in \mathbb{R}^{2N\cdot K} \label{eq4} \\
    y^- &= \{k|z_i^{(k)} < \delta, \ \forall i \in {1, \ldots, N}\} \label{eq5} \\
    (y^+, y^-) &= \text{MoLAM}(\mathbf{h}(x)) \label{eq6}
\end{align}

\subsection{Robust Active Learning}
With MoLAM, the labels generated by a Mixture of LLMs become more reliable than single LLM. However, due to the inherent noise in LLM-generated annotations, we introduce robust active learning, which leverages implicit information embedded in both the LLMs outputs and the AL model predictions to further guide the training.

Specifically, we employ negative labels $y^-$ as the set of classes that all LLMs assign consistently low confidence to, as defined in Eq.~\eqref{eq5}. To discourage the AL model from predicting these likely incorrect labels, we incorporate a negative learning loss $\mathcal{L}_{neg}$ into the training objective. This loss penalizes the model for assigning high probability to any class in $y^-$, and can be formulated as:
\begin{align}
    \mathcal{L}_{neg}= \textstyle -\sum_{k\in y^-}\log(1-p(k|x)) \label{eq7}
\end{align}
where $p(k|x)$ is the predicted probability of class $k$ by the AL model for input $x$, and $y^-\subset \{1, \ldots,K\}$ is the set of negative labels provided by MoLAM.

Moreover, the AL model can be regarded as task-specific SLM, trained for a particular task and thus more likely to encode domain-relevant knowledge. We apply the disagreement between the AL model predicted label and the MoLAM generated label as an indicator of annotation discrepancy, denoted as $d_{anno}$, and formularized as $d_{anno} = \mathbb{I}[\tilde{y} \neq y^+]$, where $\tilde{y}=\text{argmax}_k p(k|x)$, denotes the predicted class by the AL model. Empirically, we observe that $d_{anno}$ is effective in identifying erroneous labels produced by LLMs.
To avoid training leakage, we compute $d_{anno}$ using the AL model's predictions from the previous iteration, before the newly selected samples have been incorporated into training. A detailed analysis of the effectiveness of $d_{anno}$ is presented in the Component Effectiveness Analysis section under Experiments.
We incorporate $d_{anno}$ as a weight to emphasize high confidence annotations during training. We define the sample specific weight $\mathcal{W}_d$ as:
\begin{align}
    \mathcal{W}_d &= 
        \left\{ \begin{array}{cc}
            1 & if \ \ d_{anno}=0 \\
            \alpha & if \ \  d_{anno}=1 
        \end{array} \right. \label{eq8}
\end{align}
where $\alpha\in(0,1)$ is a down-weighting factor that reduces the influence of potentially incorrect annotations.
The final loss function for the AL model is then defined as:
\begin{align}
    \mathcal{L} &= \mathcal{W}_d \cdot \mathcal{L}_{CE} + \lambda\mathcal{L}_{neg} \label{eq9}
\end{align}
where $\mathcal{L}_{CE}$ is the standard cross-entropy loss for multi-class classification, $\mathcal{L}_{neg}$ is the negative learning loss, and $\lambda$ is hyperparameter controlling the weight of the negative learning penalty term. The complete training and update strategy of the MoLLIA framework is summarized in Algorithm~\ref{alg:algorithm}.

\begin{algorithm} [!t]
    \caption{MoLLIA Training and Update Strategy}
    \label{alg:algorithm}
    \textbf{Input}: Labeled pool $L$; Unlabeled pool $U$; annotation model MoLAM; AL model; query size $B$; annotation discrepancy $d_{anno}$; negative labels $y^-$.\\
    \textbf{Output}: Updated labeled and unlabeled pool, AL model, annotation discrepancy, negative labels.\\
    \begin{algorithmic}[1]
        \FOR{AL iteration}
            \STATE Select a batch of $B$ instances $x = \{x_1, x_2, \ldots, x_B\}$ from $U$ using the query strategy
            \STATE Obtain MoLAM-generated labels $y^+$ and negative labels $y^-$ of $x$ from MoLAM via Eq.~\eqref{eq6}
            \STATE Update Labeled and unlabeled pool: \\$L \leftarrow L+\{(x,y^+)\}$; $U \leftarrow U-\{x\}$
            \STATE Obtain AL-predicted label $\tilde{y} = \arg\max_k p(k|x)$
            \STATE Compute negative learning loss $\mathcal{L}_{neg}$ via Eq.~\eqref{eq7}
            \STATE Compute annotation discrepancy $d_{anno} = \mathbb{I}[\tilde{y} \neq y^+]$
            \STATE Compute weight $\mathcal{W}_d$ via Eq.~\eqref{eq8}
            \STATE Update the AL model via Eq.~\eqref{eq9}
        \ENDFOR
    \end{algorithmic}
\end{algorithm}

\section{Experiments}
To evaluate the performance and robustness of our proposed framework, we use four benchmark multi-class text classification datasets that are widely adopted in active learning research via Hugging Face platform\footnote{\url{https://huggingface.co/datasets}}: AG News~\cite{zhang2015character}, IMDB~\cite{maas2011learning}, TREC~\cite{li2002learning}, and PubMed~\cite{dernoncourt2017pubmed}.
AG News is a news classification dataset composed of news titles and descriptions; IMDB is a collection of movie reviews for sentiment classification; TREC is a question classification dataset contains open-domain, fact-based questions; PubMed is a biomedical text classification dataset composed of article abstracts focused on diabetes-related topics. Table~\ref{tab:data-exp} provides a detailed summary of these datasets. The train, validation, and test splits used in our experiments follow the original dataset configurations.

\begin{table}
    \centering
    \adjustbox{max width=\linewidth}{
    \begin{tabular}{lcccc}
        \toprule
        \multirow{2}{*}{\bfseries Dataset} & \bfseries \#Vocab./ & \multicolumn{3}{c}{\bfseries \#Document} \\
         & \bfseries \#Label & \bfseries Train & \bfseries Vali. & \bfseries Test \\
        \midrule
        AG News & 65,043/4 & 114,000 & 6,000 & 7,600 \\
        IMDB & 74,891/2 & 22,500 & 2,500 & 25,000 \\
        TREC & 8,446/6 & 5,000 & 452 & 500 \\
        PubMed & 45,457/5 & 176,642 & 29,672 & 29,578 \\
        \bottomrule
    \end{tabular}}
    \caption{Experiment used dataset statistics.}
    \label{tab:data-exp}
\end{table}

\subsection{Implementation}
To balance both performance and deployability, we adopt five widely used lightweight LLMs, each ranging from 7B to 9B parameters—suitable for inference on a single GPU with 24GB VRAM. The selected models include Gemma-2-9B-it, Llama-3.1-8B-Instruct, Mistral-7B-Instruct-v0.2, Qwen2.5-Coder-7B-Instruct, and Yi-1.5-9B. And the prompt is shown in Appendix.

We employed two pretrained language models as backbone classifiers, DistilBERT \citep{sanh2019distilbert} and DistilRoBERTa \citep{liu2019roberta}, implemented using PyTorch \cite{paszke2019pytorch}. To better simulate real-world deployment scenarios, we applied the cold start strategy \citep{zhu2019addressing} with random initialization at the beginning of each active learning iteration \citep{frankle2018lottery}. 
All experiments were conducted on a single NVIDIA A40 GPU. The maximum input sequence length was set to 128 tokens, and each training iteration was run for up to 40 epochs. The initial labeled training set and query batch size were both set to 50 instances. To prevent overfitting and improve training efficiency, we applied early stopping with a patience of 10 epochs \citep{du2019gradient, ying2019overview}.
We used the AdamW optimizer \citep{loshchilovdecoupled}, and the learning rate was set to 5e-5.

The annotation model, MoLAM, is implemented with XGBoost \cite{chen2016xgboost}. It is trained on 50 instances, identical to the initial labeled training set, randomly selected from the training set and validated on the corresponding validation set. The thresholds for pseudo-labeling and negative label identification are set to 0.9 and 0.001, respectively. The annotation discrepancy weight $\alpha$ fixed at 0.5, while the negative learning weight $\lambda$ increases linearly from 0.4 to 1 during AL iteration. The XGBoost hyperparameter for each dataset are provided in Appendix.

\subsection{Baselines}
As the field of LLMs in the Loop active learning is still in its early stages and prior work primarily adopts either a single LLM as the annotator or relies on human–LLM hybrid setups, there is currently no established baseline for multi-LLMs annotation frameworks. Therefore, to provide a meaningful comparison, we evaluate our proposed MoLLIA framework against single-LLM annotation across four widely used active learning query strategies.
To demonstrate the generalization capability of our framework, we adopt representative query strategies from three major categories: uncertainty-based, diversity-based, and hybrid approaches. 
NoiseStability \cite{li2024deep}, an uncertainty-based strategy, selects instances based on the variability of model predictions under perturbations. 
CoreSet \cite{sener2018active}, a diversity-based strategy, identifies a subset of samples that best represents the entire unlabeled pool by maximizing coverage in the feature space.
BEMPS \cite{tan2023bayesian}, a hybrid strategy, computes a proper scoring rule for each instance based on the model’s predictive distribution, effectively capturing both uncertainty and representativeness.
In addition, we include Random Sampling as a baseline to serve as a reference point for performance without active query selection.

To evaluate the effectiveness of our annotation module, MoLAM, we compare it against a diverse set of baselines that share the common objective of enhancing annotation quality. These include approaches based on LLM-ensemble, data augmentation, meta-learning, and semi-supervised learning. Since the classification task does not involve semantically meaningful output text, we adopt two output-based LLM-ensemble methods as baselines: vote-based \cite{li2024more} and logits-based \cite{fathullah2023logit}. These methods aggregate the predicted labels or the predicted probability distributions from multiple LLMs to produce the final annotation. 
To address the challenge of limited labeled data, we additionally consider vocabulary-level data augmentation (DA) \cite{ma2019nlpaug} and sentence paraphrasing (PAG) \cite{yadav2024pag} as baselines, both aimed at enriching the input space.
Furthermore, since the Mixture of LLMs paradigm can also be viewed as a form of black box meta-learning, we compare MoLAM with SNAIL \cite{mishra2018simple}, a representative meta-learning method designed to adapt rapidly from few-shot examples.
To assess the ability of MoLAM to utilize unlabeled data, we benchmark MoLAM against FixMatch \cite{sohn2020fixmatch}, a widely adopted semi-supervised learning method. 
Lastly, we include MoL, an ablation variant of MoLAM that excludes the pseudo-labeling mechanism, to isolate its contribution to overall performance.

\begin{table}
    \centering
    \adjustbox{max width=\linewidth}{
    \begin{tabular}{@{\hskip 5pt}l@{\hskip 5pt}l@{\hskip 5pt}c@{\hskip 5pt}c@{\hskip 5pt}c@{\hskip 5pt}c@{\hskip 5pt}}
        \toprule
         & \bfseries Methods & \bfseries AG News & \bfseries IMDB & \bfseries TREC & \bfseries PubMed\\
        \midrule
        \multirow{5}{*}{\makecell[l]{\bfseries Single\\ \bfseries LLM}} & GEMMA & 0.8349 & 0.9373 & 0.5741 & 0.7313 \\
        & LLAMA & 0.7908 & 0.9172 & 0.4870 & 0.6233 \\
        & MISTRAL & 0.8182 & 0.8835 & 0.6357 & 0.6257 \\
        & QWEN  & 0.7763 & 0.9403 & 0.6566 & 0.6217 \\
        & YI  & 0.7930 & 0.9503 & 0.7682 & 0.6755 \\
        \midrule
        \multirow{2}{*}{\makecell[l]{\bfseries LLM\\ \bfseries Ensem.}} & Vote-based & 0.8247 & 0.9418 & 0.7320 & 0.6802 \\
        & Logits-based & 0.8262 & 0.9421 & 0.7113 & 0.6802 \\
        \midrule
        \multirow{5}{*}{\bfseries Others}& DA & 0.5487 & 0.9206 & 0.4186 & 0.6864 \\
        & PAG & 0.7771 & 0.9322 & 0.7152 & 0.7444 \\
        & SNAIL & 0.8342 & 0.9489 & 0.6903 & 0.7199 \\
        & FixMatch & 0.8530 & 0.9490 & 0.7207 & 0.7096 \\
        & MoL & 0.8819 & 0.9534 & 0.7924 & 0.7744 \\
        & MoLAM & \bfseries 0.8887 & \bfseries 0.9538 & \bfseries 0.8040 & \bfseries 0.7772 \\
        \bottomrule
    \end{tabular}}
    \caption{Annotation accuracy across datasets for different approaches.}
    \label{tab:molam}
\end{table}

\subsection{Annotation Performance of MoLAM}
Table.~\ref{tab:molam} presents the annotation performance of our proposed MoLAM compared with other annotation baselines, evaluated on a randomly sampled test set. For the first two sections, Single LLM and LLM-ensemble, the reported accuracy is obtained by directly applying each method to label the input without any additional training. For the remaining methods involve training-based approaches, they are trained on 50 instances, same as the size of initial labeled pool. To maintain the consistency, the model backbones used for DA and PAG are XGBoost, the same as MoLAM. While for SNAIL and FixMatch, we adopt a MLP backbone with residual connections, in line with the architectural requirements of these methods, as XGBoost does not support gradient-based optimization.


\begin{figure}[!t]
    \centering
    \includegraphics[width=\columnwidth]{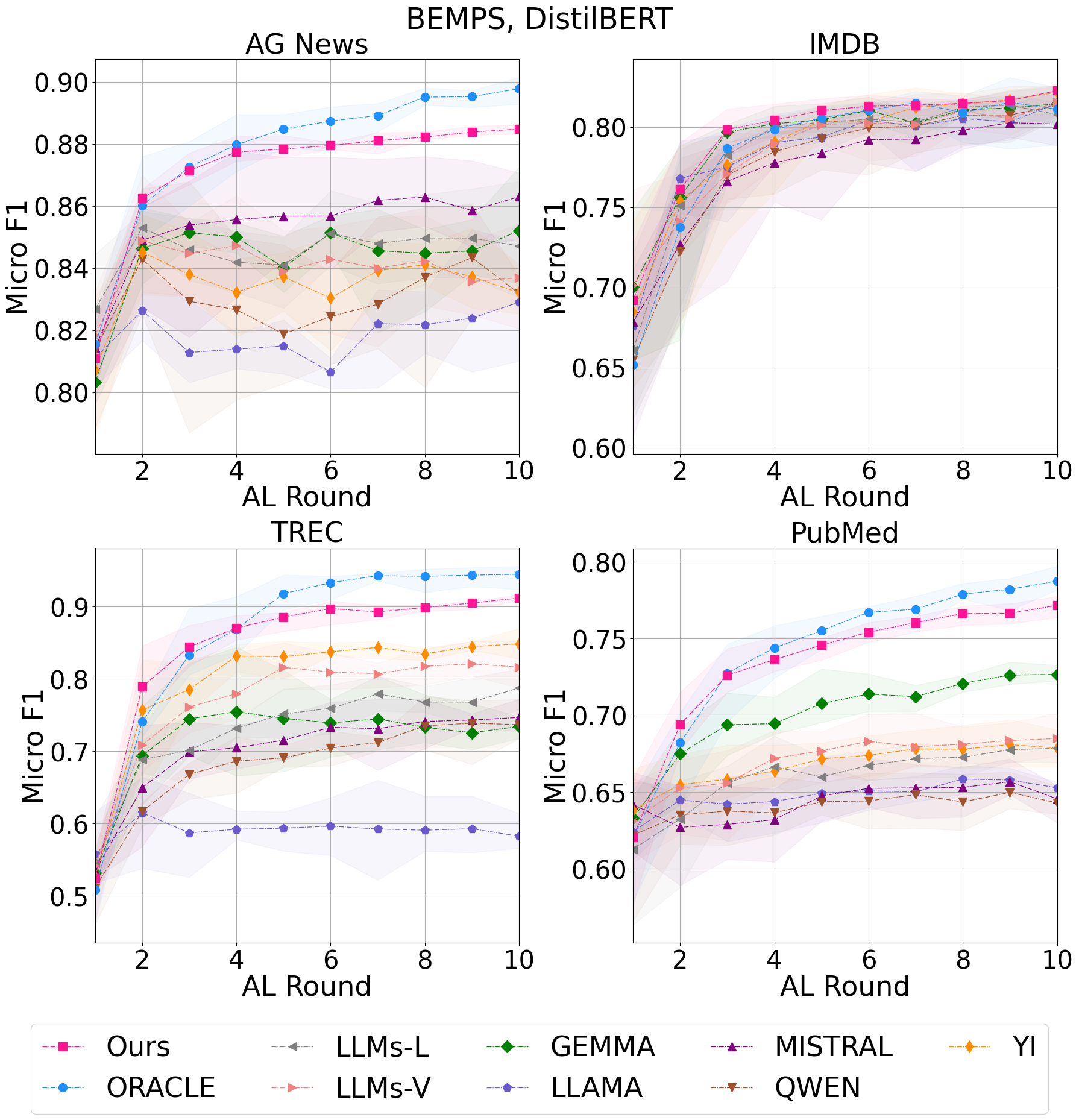} 
    \caption{Averaged micro-F1 score with BEMPS on DistilBERT, averaged results with 5 random seeds.}
    \label{fig:res1}
\end{figure}

From the table, we first observe that individual LLMs exhibit varied annotation performance across different datasets, and no single model consistently outperforms on all four benchmarks. This highlights the necessity of a Mixture-of-LLMs-based annotation model to improve generalization and robustness across diverse datasets. Additionally, we find that IMDB appears to be relatively easier for LLMs to annotate, whereas TREC and PubMed present greater challenges. This may be attributed to differences in the semantic meaning and complexity of the labels inherent to each dataset.

Among the methods that aggregate outputs from multiple LLMs, MoLAM consistently achieves the highest annotation performance across all datasets and baselines. While ensemble-based approaches help stabilize predictions by combining outputs from multiple LLMs, they do not lead to substantial improvement in annotation quality. Input data augmentation methods, DA and PAG, also exhibit limited effectiveness, likely because both vocabulary-level and sentence-level augmentations fail to enrich the input feature space; in particular, word-level transformations may distort the original semantics, resulting in degraded annotation performance.
While advanced trainable methods, SNAIL and FixMatch, employ more complex architectures with attention mechanisms, their performance still falls short of MoLAM. This could be attributed to the limited expressiveness of the input features, which constrains the effectiveness of sophisticated architectures in this annotation setting.
Moreover, the usage of pseudo-labeling in MoLAM plays a key role in effectively leveraging unlabeled data, further enhancing its annotation performance.
We also present specific examples of labels generated by MoLAM compared with those from ensemble methods, as shown in Appendix Fig.~12.

\subsection{Active Learning Performance}
\begin{figure}
    \centering
    \includegraphics[width=\columnwidth]{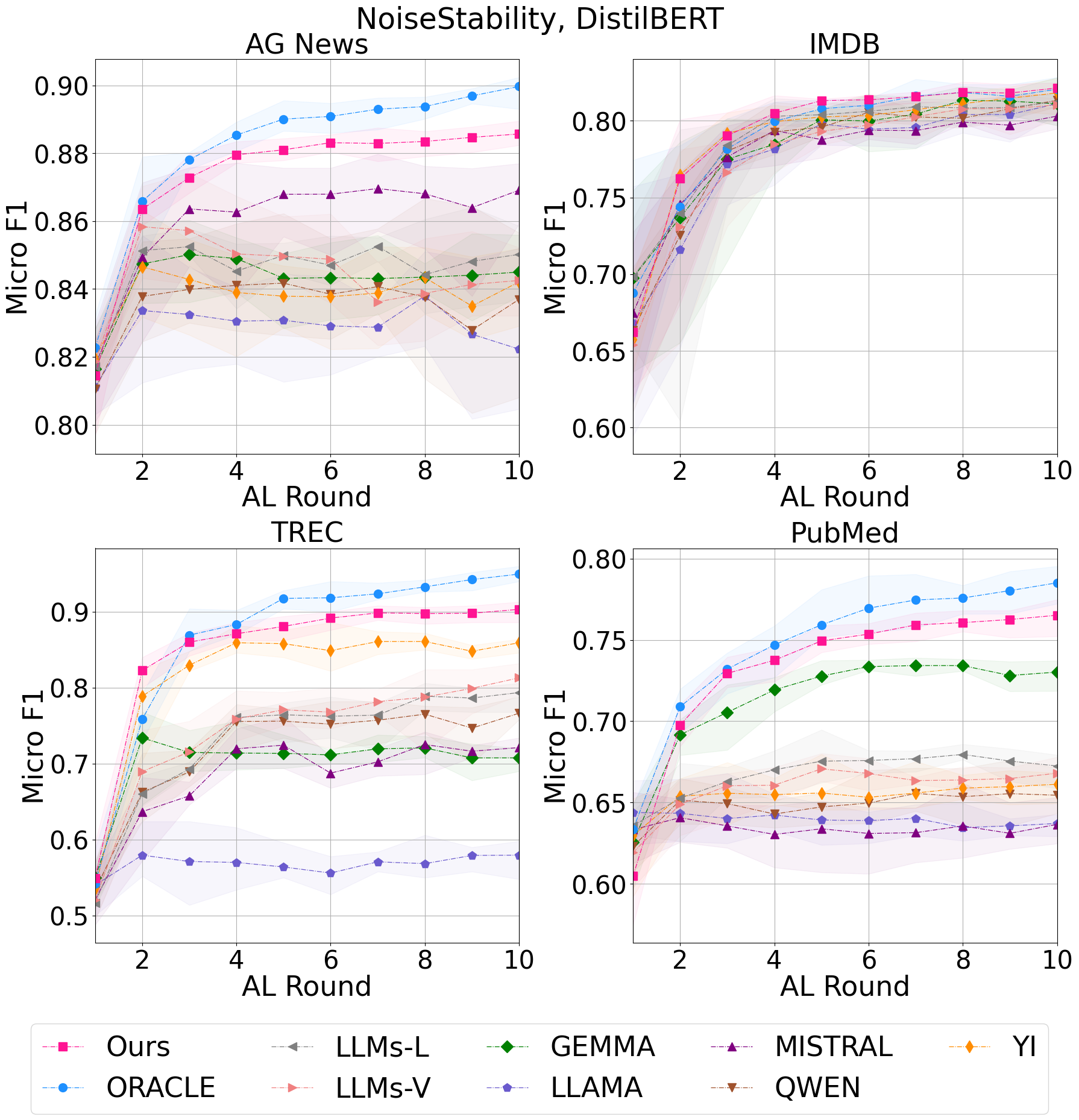} 
    \caption{Averaged micro-F1 score with NoiseStability on DistilBERT, averaged results with 5 random seeds.}
    \label{fig:res2}
\end{figure}

Fig.~\ref{fig:res1} and \ref{fig:res2}, along with Fig.~\ref{fig:res3} and \ref{fig:res4} in the Appendix, illustrate the performance of our proposed framework across different datasets, backbone classifiers, and query strategies. LLMs-L and LLMs-V refer to the LLM-ensemble methods based on logits and voting, respectively.
Overall, we observe that MoLLIA consistently outperforms both individual LLM and ensemble-based methods, achieving performance comparable to that of human annotators. Notably, on simpler datasets such as IMDB, which contains only two labels, MoLLIA even surpasses the oracle annotations. This is attributed to the incorporation of negative learning and annotation discrepancy, which enhance learning effectiveness.
For datasets that are more challenging for LLMs to annotate, MoLLIA still demonstrates superior performance over both single LLM and ensemble methods, closely matches human-level annotation quality, and highlighting its potential as a practical substitute for human annotators. 
While some single LLM, e.g. Llama and Yi, perform well on specific datasets, their performance is inconsistent and tends to drop significantly on others, highlighting the lack of generalization across diverse tasks.

\subsection{Component Effective Analysis and Ablation Study}
\begin{table}
    \centering
    \adjustbox{max width=\linewidth}{
    \begin{tabular}{@{\hskip 5pt}l@{\hskip 5pt}c@{\hskip 5pt}c@{\hskip 5pt}c@{\hskip 5pt}c@{\hskip 5pt}}
        \toprule
         & \textbf{AG News} & \textbf{IMDB} & \textbf{TREC} & \textbf{PubMed} \\
        \midrule
        True Negative Labels & 0.4833 & 0.2142 & 0.2511 & 0.1844 \\
        False Negative Labels & 0.0030 & 0.0002 & 0.0001 & 0.0022 \\
        \bottomrule
    \end{tabular}}
    \caption{Distribution of negative labels across datasets, reported as the proportion relative to the full label space.}
    \label{tab:nl}
\end{table}

\begin{figure}
    \centering
    \includegraphics[width=\columnwidth]{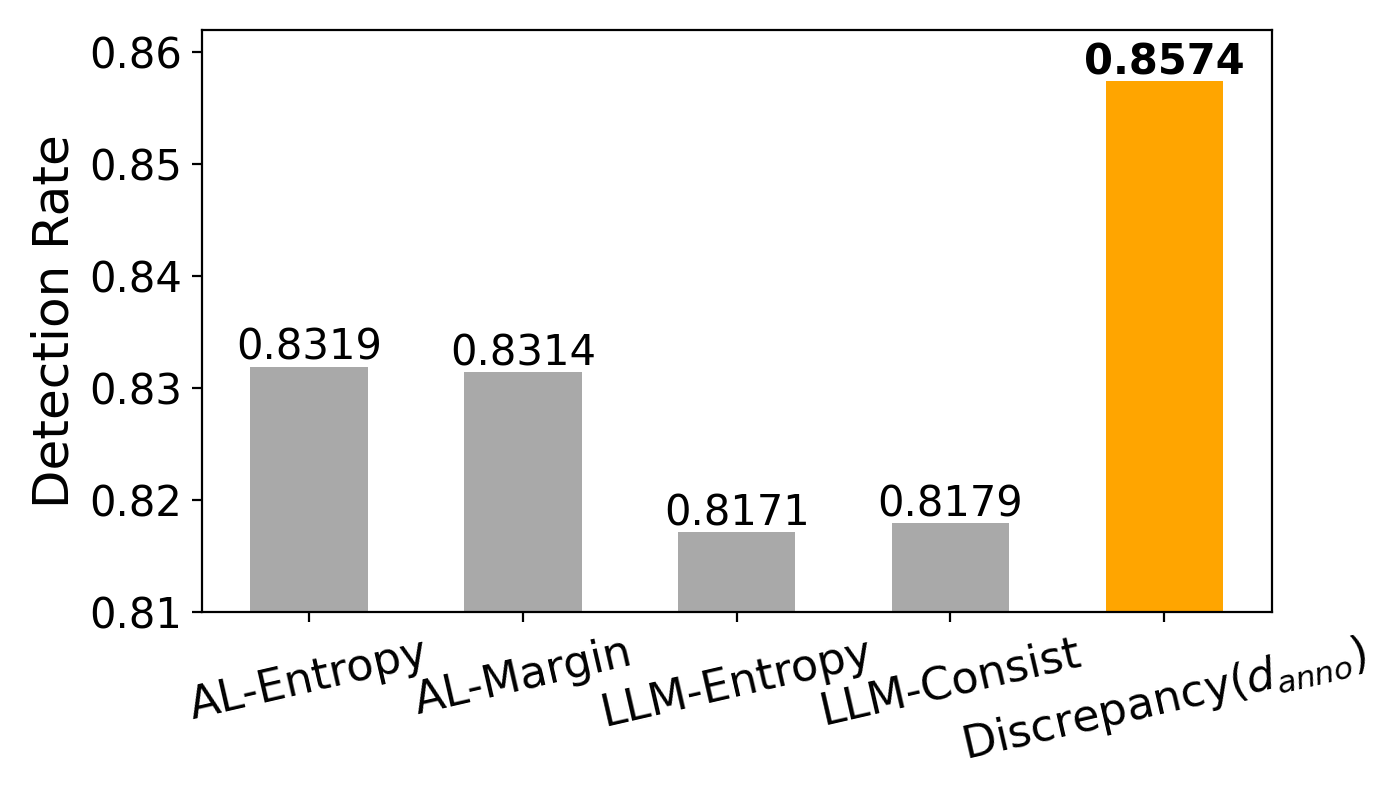} 
    \caption{True positive detection rate of annotation discrepancy across different estimation methods, measured as the proportion of correctly identified accurate annotations among all annotated instances.}
    \label{fig:d-anno}
\end{figure}

To evaluate the effectiveness of key components in our framework, we conduct a quantitative analysis of the negative labels and annotation discrepancy.
Table~\ref{tab:nl} reports the distribution of negative labels $y^-$ provided by MoLAM with the $\delta=0.001$ across the total label space.
True Negative Labels indicate cases where MoLAM correctly identifies labels that do not belong to the instance, while False Negative Labels correspond to instances where the true label is mistakenly classified as negative.
The results demonstrate that MoLAM offers insightful guidance to the AL model through its negative label predictions. Importantly, the false negative rate remains extremely low across all datasets, suggesting that the use of negative labels introduces minimal additional noise into the learning process. 
Fig.~\ref{fig:d-anno} presents the true positive detection rate of annotation discrepancy, measured as the proportion of correctly identified accurate annotations out of all annotations. We compare our approach with several mainstream uncertainty estimation methods, including entropy, margin, and consistency-based metrics. The results show that the discrepancy derived from the disagreement between AL model and LLMs achieves the highest detection performance, highlighting its effectiveness in identifying potentially unreliable annotations.

\begin{table}
    \centering
    \adjustbox{max width=\columnwidth}{
    \begin{tabular}{lcccc}
        \toprule
         & \textbf{MoLAM} & \textbf{Negative Learning} & \textbf{Annotation Discrepancy} \\
        \midrule
        A & $\surd$ & $\surd$ & $\surd$ \\
        B & $\surd$ & $\surd$ & \\
        C & $\surd$ &  & $\surd$ \\
        D & $\surd$ &  &  \\
        \bottomrule
    \end{tabular}}
    \caption{Ablation study configurations.}
    \label{tab:ablation}
\end{table}
\begin{figure}[t]
    \centering
    \includegraphics[width=\columnwidth]{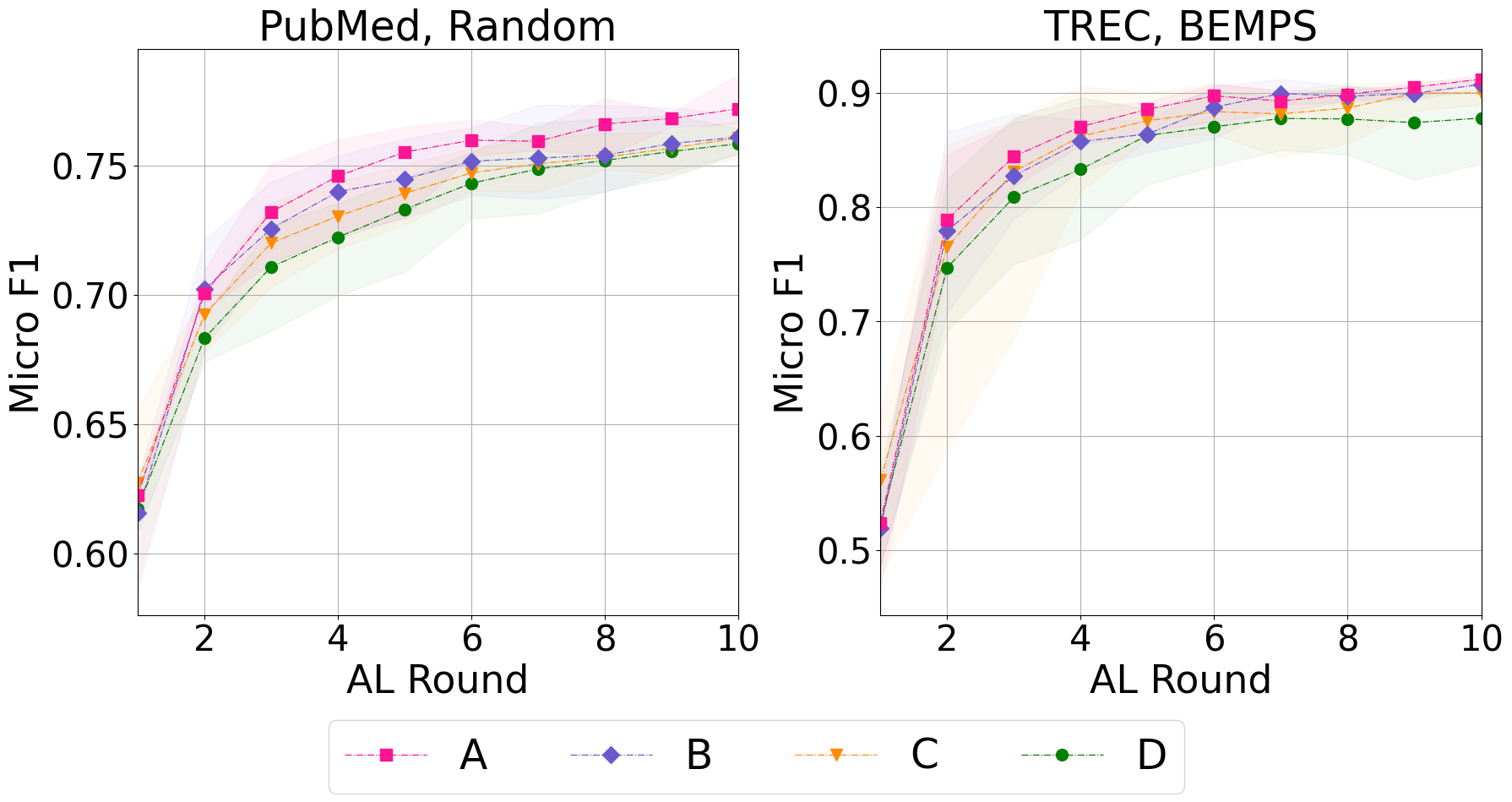} 
    \caption{Ablation study of different component, and the legend is referred from Table~\ref{tab:ablation}.}
    \label{fig:ablation}
\end{figure}

To assess the contribution of each core component in our proposed MoLLIA framework, we conduct an ablation study with four variants, as summarized in Table~\ref{tab:ablation}. Each variant disables one or more components, Negative Learning and Annotation Discrepancy, while retaining the MoLAM annotation module. The corresponding experimental results are presented in Fig.~\ref{fig:ablation}, with each curve labeled according to its configuration in the table.
The results clearly show that both Negative Learning and Annotation Discrepancy significantly contribute to the overall performance of the model. Notably, the inclusion of Negative Learning yields the most substantial performance improvement, underscoring its critical role in enhancing the learning effectiveness of the AL model model by guiding it away from wrong predictions.

Additionally, we conduct a parameter sensitivity analysis to evaluate the generalizability of MoLLIA, as shown in Appendix Fig.~10 and Fig.~11.
To demonstrate the deployability of the proposed framework, we report the maximum CUDA memory usage across different datasets and backbone models in Appendix Table~5. 

\section{Conclusion}
In this study, we proposed a novel active learning framework that replaces human annotators with a Mixture-of-LLMs-based annotator, significantly reducing annotation costs. To ensure practical applicability and robustness in real-world scenarios, our framework relies solely on lightweight LLMs and incorporates negative learning and annotation discrepancy to further enhance the learning effectiveness of the AL model.
Overall, the proposed MoLLIA framework demonstrates strong performance across four benchmark datasets, achieving annotation quality comparable to that of human annotators.
Future work may explore hybrid annotation strategies that combine LLM-generated and human labels to balance efficiency and accuracy.

\section{Acknowledgments}
This research was supported by an Australian Government Research Training Program (RTP) Scholarship.

\bibliography{aaai2026}

\appendix
\section{Appendix}
\subsection{Additional Result}
\begin{figure}[H]
    \centering
    \includegraphics[width=\columnwidth]{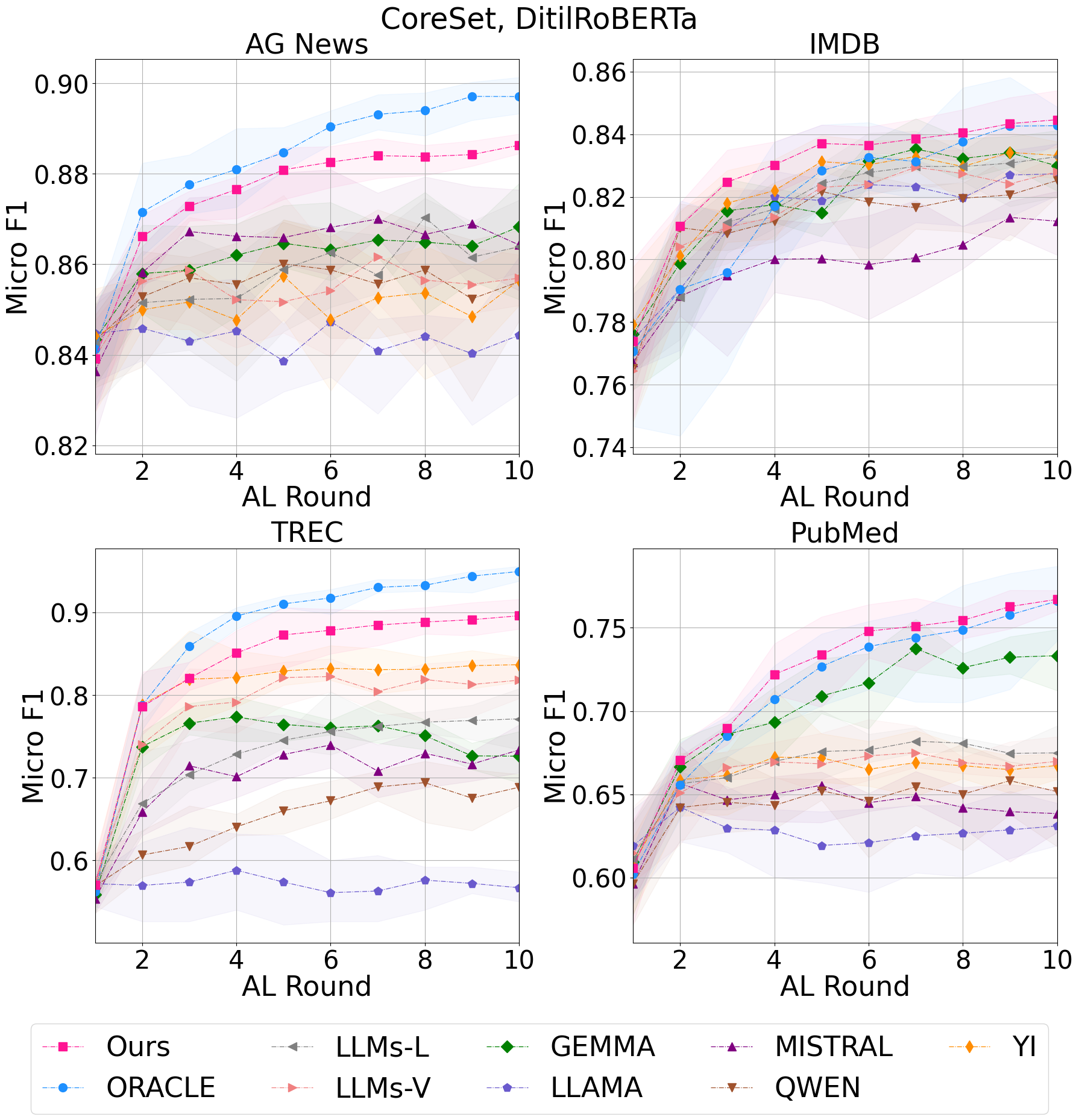} 
    \caption{Averaged micro-F1 score with CoreSet on DistilRoBERTa, averaged results with 5 random seeds.}
    \label{fig:res3}
\end{figure}

\begin{figure}[H]
    \centering
    \includegraphics[width=\columnwidth]{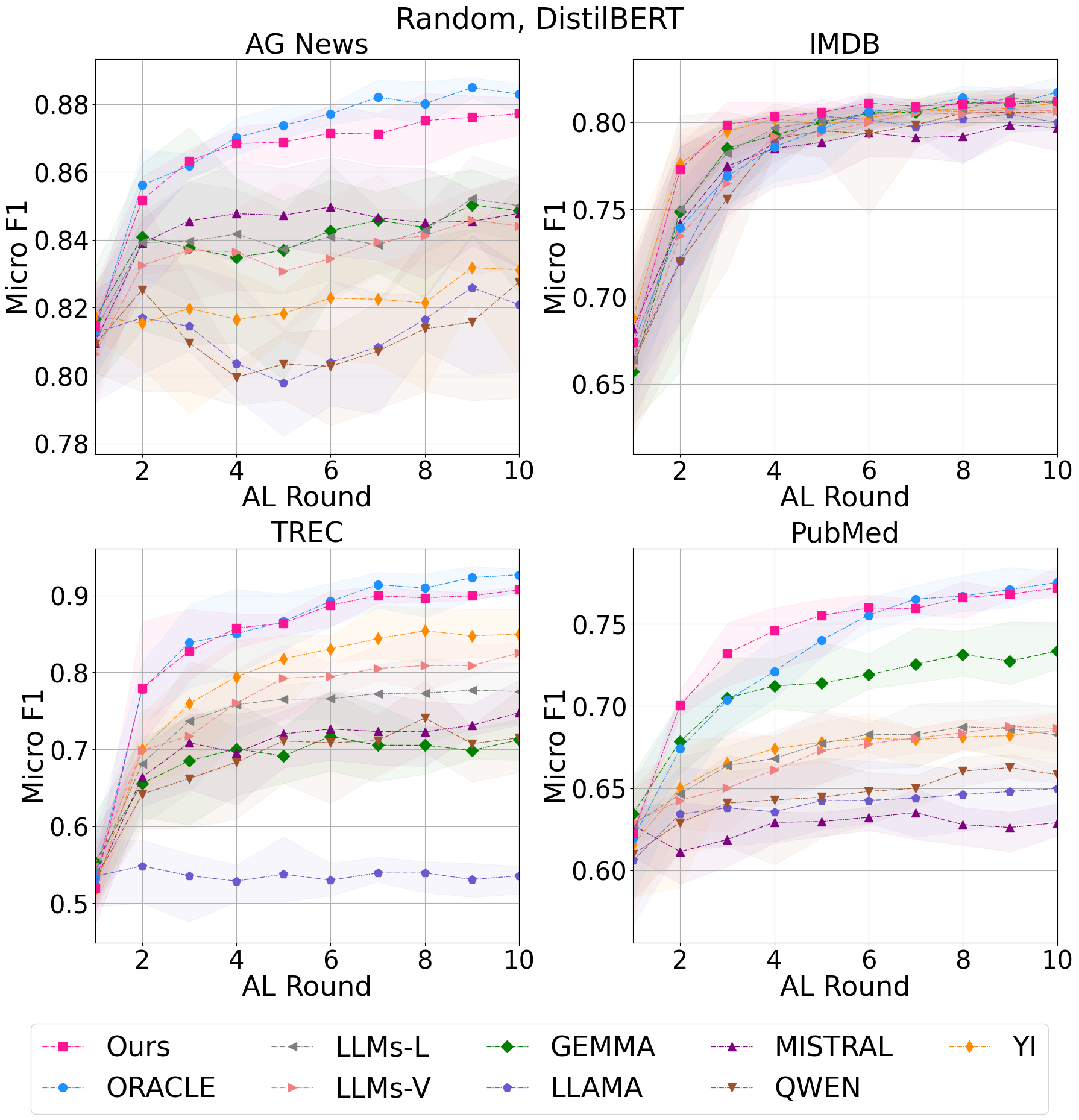} 
    \caption{Averaged micro-F1 score with random on DistilBERT, averaged results with 5 random seeds.}
    \label{fig:res4}
\end{figure}

\subsection{Parameter Sensitive Analysis}
\begin{figure}[H]
    \centering
    \includegraphics[width=\columnwidth]{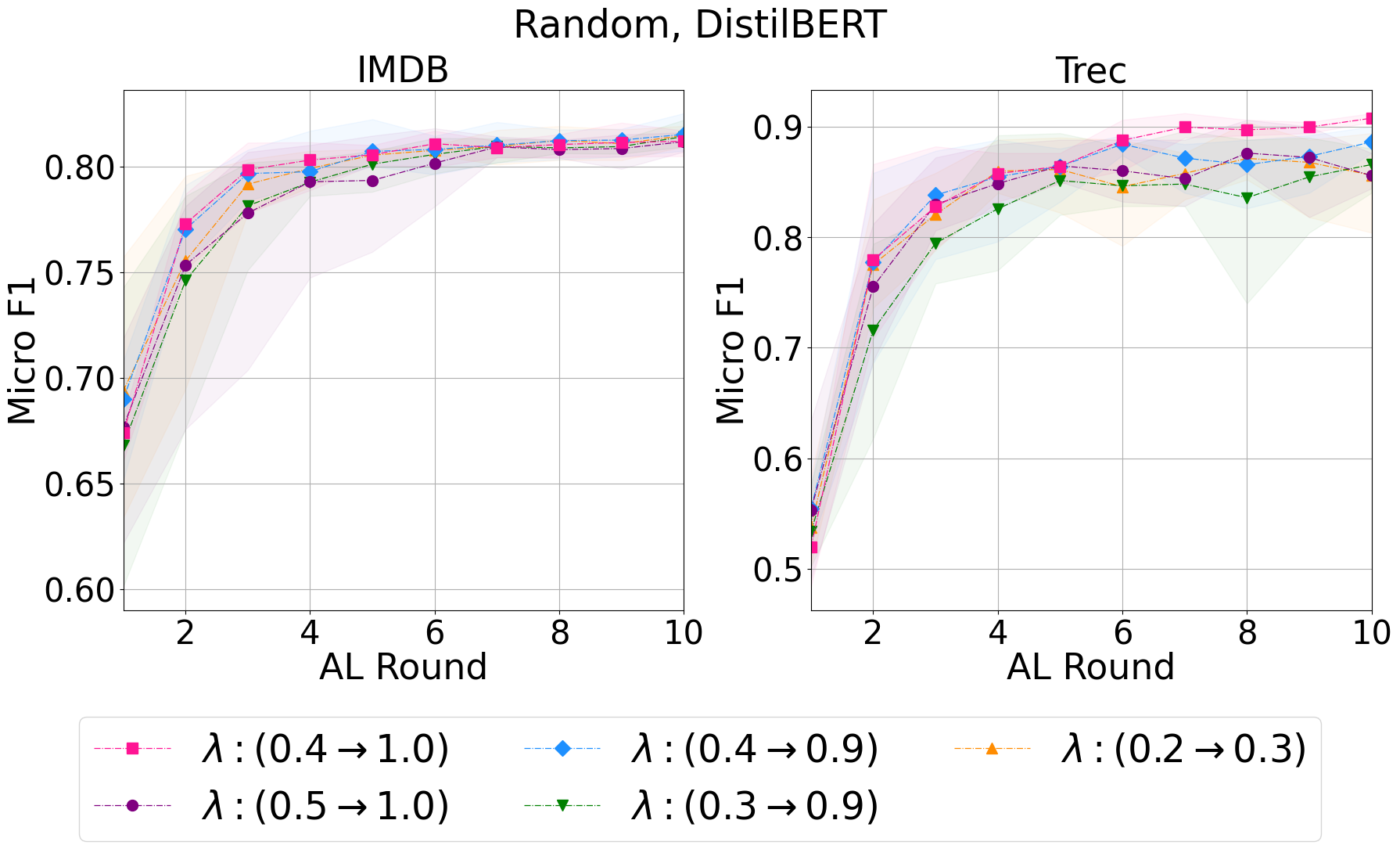} 
    \caption{Performance on different negative learning weight parameter $\lambda$.}
    \label{fig:param1}
\end{figure}

\begin{figure}[H]
    \centering
    \includegraphics[width=\columnwidth]{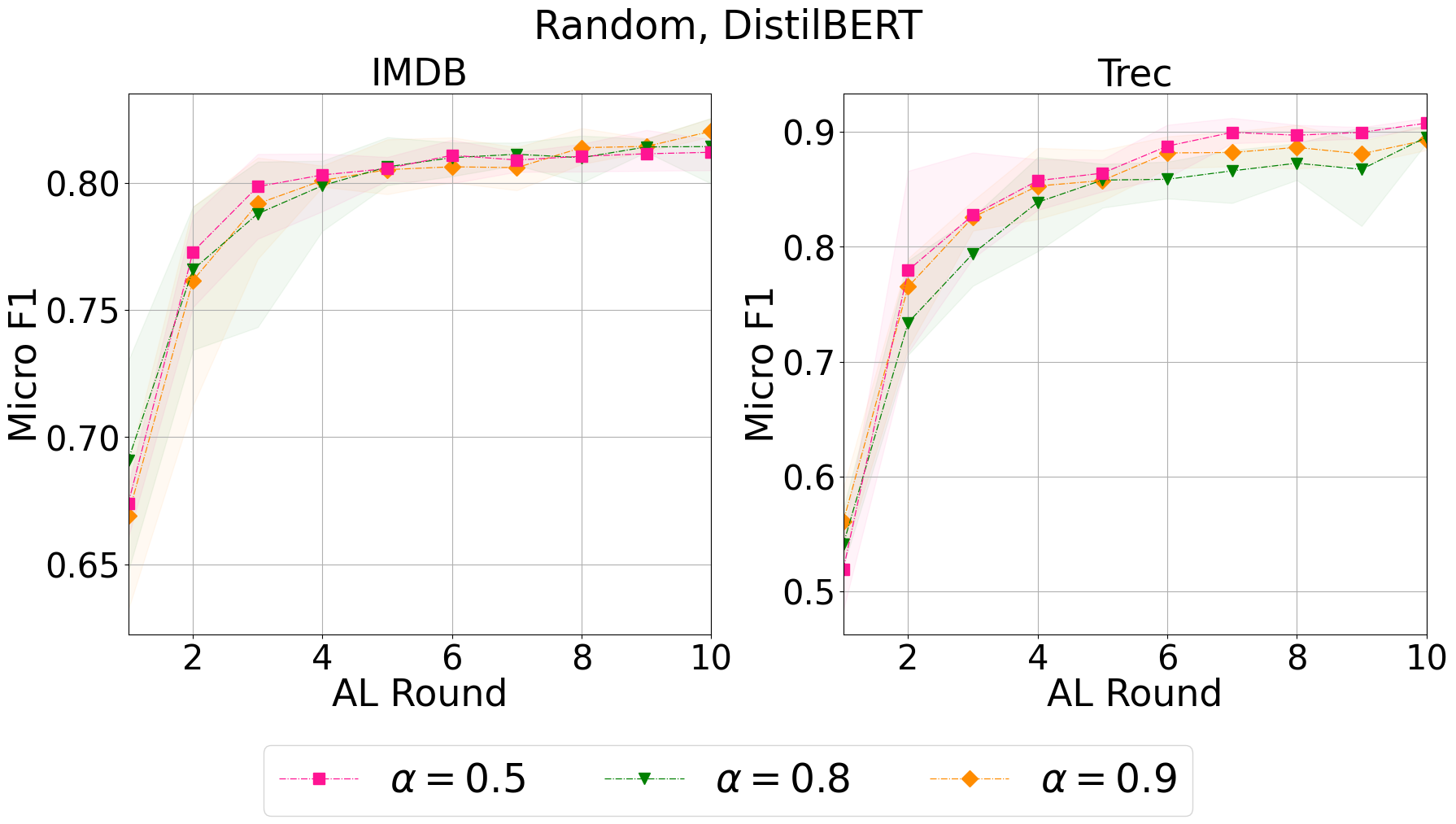} 
    \caption{Performance with different values of the annotation discrepancy weight parameter $\alpha$.}
    \label{fig:param2}
\end{figure}

Figures~\ref{fig:param1} and~\ref{fig:param2} present the parameter sensitivity analysis for the negative learning weight $\lambda$ and the annotation discrepancy weight $\alpha$. The results are reported as micro-F1 scores, averaged over five random seeds using the Random query strategy on DistilBERT. Since $\lambda$ is linearly increased during AL iterations, the legends indicate its starting and ending values. Overall, the results show that increasing $\lambda$ benefits the learning process by gradually incorporating negative labels. Additionally, setting $\alpha = 0.5$ allows the model to effectively leverage annotation discrepancies, helping it focus on more reliable supervision.

\subsection{Qualitative Evaluation of MoLAM}

\begin{figure}
    \centering
    \begin{subfigure}{\columnwidth}
        \centering
        \includegraphics[width=\columnwidth]{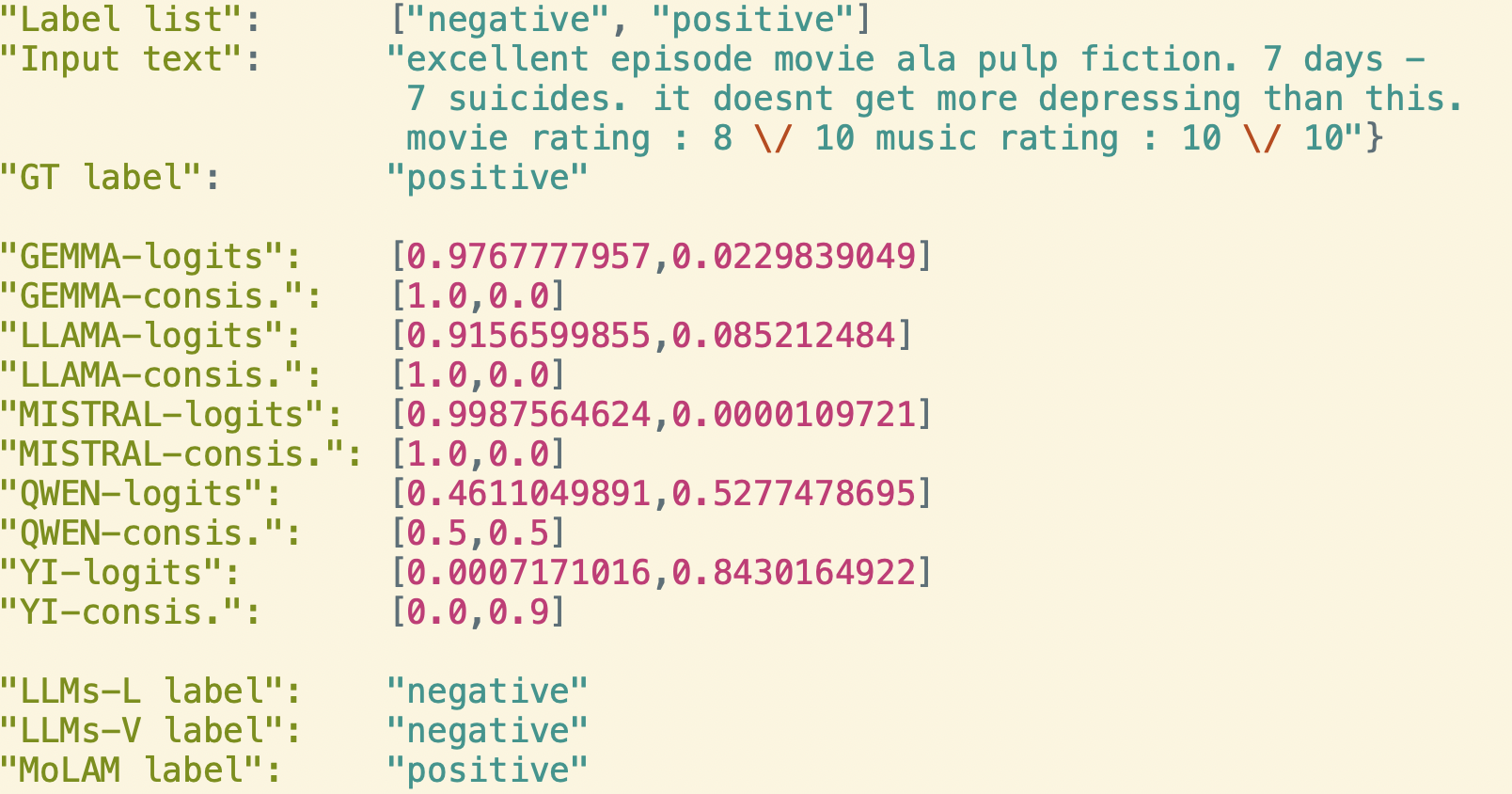}
        \caption{An example on IMDB.}
        \label{fig:ensemble}
    \end{subfigure}
    \begin{subfigure}{\columnwidth}
        \centering
        \includegraphics[width=\columnwidth]{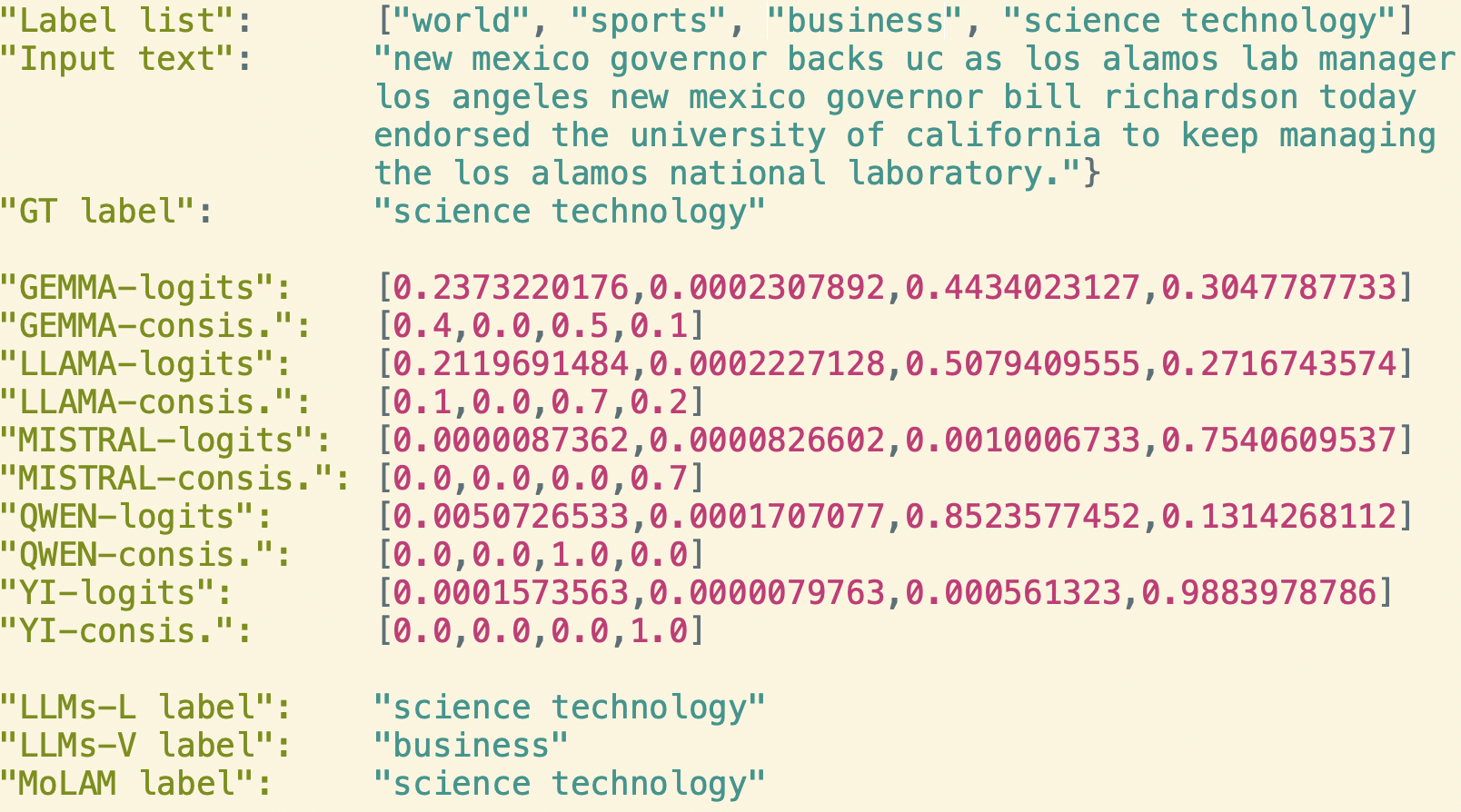}
        \caption{An example on AG News.}
        \label{fig:mixture}
    \end{subfigure}
    \caption{Qualitative comparison of annotation performance between LLM-ensemble and MoLAM.}
    \label{fig:case}
\end{figure}

Fig.~\ref{fig:case} presents two examples illustrating the annotation performance of MoLAM. In these examples, LLMs-L and LLMs-V represent the LLM-ensemble baselines based on logits aggregation and majority voting, respectively. The values shown in square brackets correspond to the logits (for LLMs-L) or consistency scores (for LLMs-V) associated with each label index in the label list. In some cases, the sum of the consistency scores does not equal 1 because certain LLMs may generate non-existent or invalid labels.

All logits and consistency scores from the participating LLMs serve as input features to MoLAM. Notably, in both examples, MoLAM successfully predicts the correct label even when both ensemble baselines fail. This highlights the effectiveness of our proposed Mixture-of-LLMs-based annotation model in capturing nuanced decision patterns beyond simple aggregation, leading to more accurate and robust annotations.

\subsection{Prompt design}
\begin{tcolorbox}[promptbox]
Classify the given question based on the following categories: 
\textcolor{red}{\bfseries \{List of labels\}} \newline
Task: Determine the most appropriate category for the question. Your response should be only one of these labels: \textcolor{red}{\bfseries \{List of labels\}}, with no additional text or explanation. \newline
Question: \textcolor{red}{\bfseries \{article\}} \newline
Output:
\end{tcolorbox}

\subsection{MoLAM parameter} \label{sec:MoLAM-param}
\begin{table}[H]
    \centering
    \adjustbox{max width=\linewidth}{
    \begin{tabular}{lcccc}
        \toprule
         & \textbf{AG News} & \textbf{IMDB} & \textbf{TREC} & \textbf{PubMed} \\
        \midrule
        Learning rate (lr) & 0.07 & 0.01 & 0.05 & 0.01 \\
        Max depth (md) & 5 & 5 & 6 & 3 \\
        \# Estimators (ne) & 300 & 300 & 300 & 500 \\
        \bottomrule
    \end{tabular}}
    \caption{XGBoost hyperparameters used in MoLAM across datasets.}
    \label{tab:MoLAM-param}
\end{table}

\subsection{CUDA Memory Usage}
\begin{table}[H]
    \centering
    \adjustbox{max width=\linewidth}{
    \begin{tabular}{lcccc}
        \toprule
         & \textbf{AG News} & \textbf{IMDB} & \textbf{TREC} & \textbf{PubMed} \\
        \midrule
        DistilBERT & 20489 & 20780 & 20348 & 20114 \\
        DistilRoBERTa & 20516 & 20604 & 19988 & 20340 \\
        \bottomrule
    \end{tabular}}
    \caption{The maximum CUDA memory occupation (MB) during the AL iteration across different query strategies.}
    \label{tab:CUDA}
\end{table}

\end{document}